\documentclass[letterpaper, 10 pt, conference]{ieeeconf}  

\IEEEoverridecommandlockouts                              

\title{\LARGE \bf
HELM: Human-Preferred Exploration with Language Models
}

\author{Shuhao Liao$^{1}$, Xuxin Lv$^{1}$, Yuhong Cao$^{2}$, Jeric Lew$^{2}$, Wenjun Wu$^{1}$, Guillaume Sartoretti$^{2}$%
\thanks{$^{1}$Hangzhou International Innovation Institute, Beihang University, China}
\thanks{$^{2}$Department of Mechanical Engineering, National University of Singapore, Singapore}
}

\usepackage{amssymb}
\usepackage{amsmath}
\usepackage{graphicx}
\usepackage{multirow}
\usepackage{booktabs}
\usepackage{tabularx}
\usepackage{subcaption}
\usepackage{cite}
\usepackage{tcolorbox}
\usepackage[ruled]{algorithm2e}

\begin{document}

\makeatletter
\let\@oldmaketitle\@maketitle
\renewcommand{\@maketitle}{\@oldmaketitle
  \begin{center}
  \vspace{1.0em}
  \captionsetup{type=figure}
  \includegraphics[width=1\textwidth]{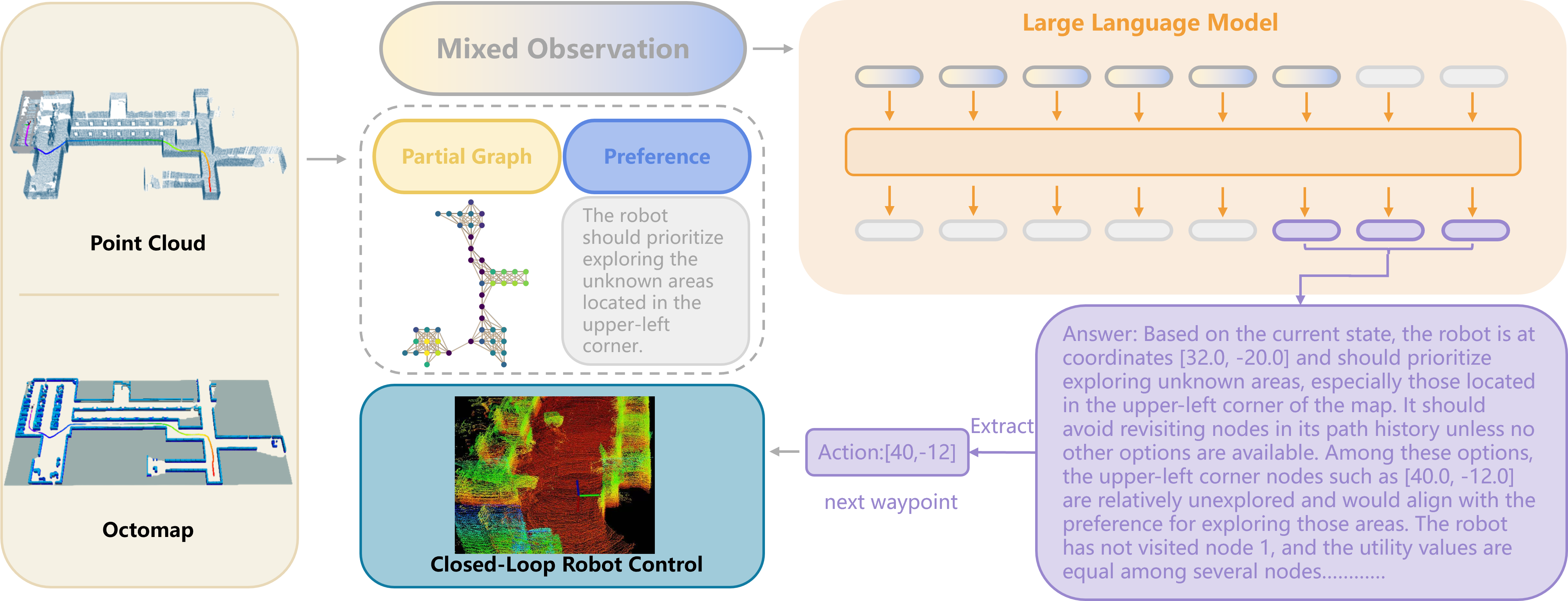}
    \captionof{figure}{Our human-preferred exploration framework uses a pre-trained large-language model (LLM), which can follow human preferences to perform long-term planning based on onboard sensor data.} 
    \label{pic1}
    \end{center}
}
\makeatother

\maketitle
\thispagestyle{empty}
\pagestyle{empty}

\begin{abstract}

In autonomous exploration tasks, robots are required to explore and map unknown environments while efficiently planning in dynamic and uncertain conditions. Given the significant variability of environments, human operators often have specific preference requirements for exploration, such as prioritizing certain areas or optimizing for different aspects of efficiency. However, existing methods struggle to accommodate these human preferences adaptively, often requiring extensive parameter tuning or network retraining. With the recent advancements in Large Language Models (LLMs), which have been widely applied to text-based planning and complex reasoning, their potential for enhancing autonomous exploration is becoming increasingly promising. Motivated by this, we propose an LLM-based human-preferred exploration framework that seamlessly integrates a mobile robot system with LLMs. By leveraging the reasoning and adaptability of LLMs, our approach enables intuitive and flexible preference control through natural language while maintaining a task success rate comparable to state-of-the-art traditional methods. Experimental results demonstrate that our framework effectively bridges the gap between human intent and policy preference in autonomous exploration, offering a more user-friendly and adaptable solution for real-world robotic applications.

\end{abstract}
\section{INTRODUCTION}

Autonomous exploration is a fundamental task in robotics where a robot is required to rapidly explore and map an unknown environment. This capability has broad applications in various domains, including search and rescue operations \cite{rescue}, environmental surveillance \cite{surveillance}, and service robotics \cite{service}. By leveraging sensors such as LiDARs or cameras, the robot gathers data about its surroundings to incrementally construct a belief of the environment, typically represented as an occupancy grid map or a voxel map. The goal of autonomous exploration is not only to achieve complete environment coverage but also to do so with optimized resource utilization, such as time and energy efficiency.
Traditional autonomous exploration approaches primarily fall into two categories: frontier-based \cite{frontier1} and sampling-based methods \cite{sample1}. Recently, these planners have been designed with delicate hierarchical structures to reduce computational complexity while maintaining the global efficiency of planned local paths \cite{tare}. On the other hand, learning-based approaches typically formulate the exploration problem as a partially observable Markov decision process (POMDP) and employ deep reinforcement learning (DRL) to derive adaptive policies by estimating long-term exploration efficiency through neural networks trained from historical experiences \cite{ariadne}. 

We note that existing methods introduce \textit{preferences} on the exploration behaviors, either from the hyperparameters in objective functions of conventional planners, or from the reward design of learning-based planners. Such preferences are usually described as the trade-off between refining explored areas and discovering new areas~\cite{ariadne}, which is the key to controlling  back-and-forth behaviours. However, preferences of existing methods are essentially non-adaptive. For conventional planners, hyperparameters are usually assumed to be fixed during the tasks. For learning-based planners, although intuitively preferences could be more adaptive, their range is severely constrained by training data, thus requiring retraining of the model to generate out-of-distribution preferences. To address this issue, the DARPA Subterranean Challenge \cite{darpa_subterranean,dapa_win2021} allowed human operators to take over the control of the robot during exploration and navigate the robot with human preference, which significantly improved the final exploration performance. That being said, incorporating those human preferences into  autonomous planners is non-trivial, as it is not straightforward to transfer human intuition to planners' objective functions.


Inspired by impressive intelligence emerging in recent development of large language models (LLMs), we propose HELM, a novel framework capable of autonomously exploring large-scale indoor environments while seamlessly integrating human preferences. HELM constructs the robot’s belief from onboard sensor data to ensure accurate environmental perception and fuses this belief with human-specified preferences to improve the adaptability of the framework. We leverage the reasoning capabilities of LLMs to generate exploration plans based on both environmental inputs and human directives. The structured output from LLM-based planning is then used for local-level motion planner, ultimately enabling closed-loop robot control. We evaluate HELM through extensive numerical comparisons against existing planners. Experimental results demonstrate that our framework achieves exploration efficiency comparable to state-of-the-art methods while providing significantly more flexible and intuitive preference control. By enabling effortless human-in-the-loop adjustments via natural language, HELM bridges the gap between autonomous decision-making and user intent, making exploration more adaptive and efficient in real-world applications.

\section{RELATED WORK}

\subsection{Autonomous Exploration}
Autonomous exploration has been widely studied using both traditional planning-based and learning-based approaches. Frontier-based and sampling-based methods typically rely on greedy strategies for short-term path planning due to the limited knowledge of the environment, often resulting in myopic exploration \cite{1997frontier,2011frontier,rttsample}. To address this, Cao et al. proposed \cite{tare}, which optimizes the full exploration path by leveraging the latest partial map, significantly improving performance in large-scale 3D environments. Similar strategies have been explored in informative path planning, which emphasizes maximizing information gain, particularly in obstacle-free settings. On the other hand, learning-based methods, primarily DRL, aim to optimize long-term exploration by training policies to maximize future rewards. While DRL-based planners often use convolutional neural networks (CNNs) to select waypoints from a visual representation of the environment, their performance remains comparable to naïve frontier-based methods in small-scale settings \cite{cnn1,cnn2,cnn3}. To enhance learning-based exploration, Chen et al. proposed a graph neural network (GNN) approach to mitigate localization errors \cite{gnn1}, though it is restricted to obstacle-free environments. Our prior work \cite{ariadne,dle} introduced learned attention over a graph representation of the robot’s belief, demonstrating superior performance over both frontier-based planners and CNN-based DRL methods.

\subsection{Large Language Models}
LLMs can handle a wide range of reasoning tasks by using text prompts to represent task-specific needs within a unified framework. This approach enables strong generalization with minimal domain-specific tuning \cite{LLMGNNAdapter}. Recent advancements have expanded LLMs' functionalities beyond traditional reasoning tasks \cite{OptiMUS,HierarchicalLanguageAgent,rstar}. For instance, the LLaGA effectively integrates LLM capabilities to handle the complexities of graph-structured data \cite{LLaGA}. LLaGA reorganizes graph nodes into structure-aware sequences and maps them into the token embedding space through a versatile projector, enabling the model to perform consistently across different datasets and tasks without task-specific adjustments. Similarly, GraphWiz is an instruction-following language model designed to tackle a broad spectrum of graph problems \cite{GraphWiz}. By utilizing a comprehensive instruction-tuning dataset, GraphWiz is capable of resolving various graph problem types while generating clear reasoning processes, surpassing models like GPT-4 in accuracy across multiple tasks. Moreover, the emergence of models like o3 and DeepSeek-R1 signify a shift towards enhancing reasoning abilities within LLMs \cite{deepseekr1}. These models are designed to perform complex reasoning tasks by generating detailed, step-by-step solutions, akin to human problem-solving processes. Such advancements highlight the potential of LLMs to tackle intricate challenges across various domains, including mathematics, science, and coding. 

\begin{figure*}[t]
  \centering
  \includegraphics[scale=0.4]{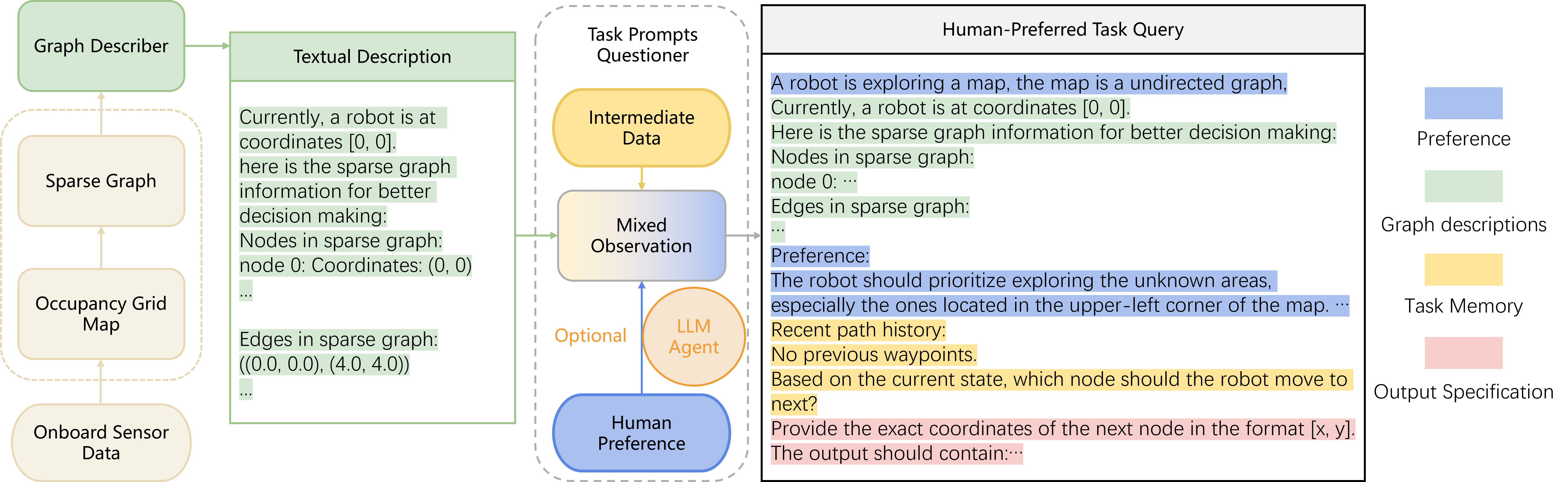}
  \caption{HELM processes onboard sensor data and combines it with human preferences and real-time intermediate data to construct human-preferred queries for decision-making. The Task Prompts Questioner can either use a predefined template or the LLM Agent for automatic generation.}
  \label{prompt}
\end{figure*}

\section{PROBLEM FORMULATION}

We consider a bounded and unknown environment $\mathcal{E}$, represented by a $x \times y$ 2D occupancy grid map. The robot incrementally constructs a partial map $\mathcal{M}$, which consists of both known and unknown regions. Specifically, we denote the unknown (unexplored) region as $\mathcal{M}_u$ and the known (explored) region as $\mathcal{M}_k$, such that:  

\begin{equation}
\mathcal{M} = \mathcal{M}_u \cup \mathcal{M}_k
\end{equation}

The known region $\mathcal{M}_k$ is further classified into:
\begin{itemize}
    \item \textbf{Free space} ($\mathcal{M}_f$): Traversable areas where the robot can navigate.
    \item \textbf{Occupied space} ($\mathcal{M}_o$): Obstacles that restrict movement.
\end{itemize}

Thus, we define:

\begin{equation}
\mathcal{M}_k = \mathcal{M}_f \cup \mathcal{M}_o
\end{equation}

At the start of exploration, the environment is entirely unknown, i.e., $\mathcal{M} = \mathcal{M}_u$. As the robot explores, regions within its sensor range $d_s$ (assuming a 360-degree LiDAR) are classified as either free space or occupied space based on sensor measurements.

The objective of autonomous exploration is to compute the shortest collision-free trajectory $\tau^*$ that fully explores the environment, such that:

\begin{equation}
\tau^* = \underset{\tau \in T}{\operatorname{argmin}} C(\tau), \quad \text{s.t. } \mathcal{M}_k = \mathcal{M}_g
\end{equation}

where:
\begin{itemize}
    \item $C: \tau \rightarrow \mathbb{R}^{+}$ maps a trajectory to its total path length.
    \item $\mathcal{M}_g$ represents the ground truth map of the environment.
\end{itemize}

Although $\mathcal{M}_g$ is not accessible during real-world exploration, it is available for evaluation in simulation or benchmark testing. In practice, most existing methods approximate exploration completion by considering the closure of occupied areas, i.e., $\mathcal{M}_k = \mathcal{M}_g$, as a stopping criterion.

\section{HUMAN PREFERRED EXPLORATION}

Our method starts from onboard sensor data and gradually constructs human-preferred queries for decision-making, completing closed-loop robot control. The construction of human-preferred queries is illustrated in Fig.\ref{prompt}.

\subsection{Building Robot Belief}

To construct the robot belief based on the environment and task requirements, we first generate a collision-free graph from onboard sensor data and then apply sparsification to optimize its structure. This graph serves as the foundation for trajectory planning. Specifically, we define the robot trajectory \(\tau\) as a sequence of viewpoints, \(\tau = (\tau_0, \tau_1, \dots)\), where each waypoint \(\tau_i \in \mathcal{M}_f\) lies within the free space.  At each decision step \(t\), we generate a set of candidate viewpoints \(V_t = \{v_0, v_1, \dots\}\), where each \(v_i = (x_i, y_i) \in \mathcal{M}_f\), uniformly distributed over the current free area \(\mathcal{M}_f\), following the approach in Tare. To maintain connectivity, we link each viewpoint to its \(k\) nearest neighbors through direct edges and remove connections that intersect occupied or unknown areas. The resulting sparsified collision-free graph \(G_t = (V_t, E_t)\) consists of the optimized set of viewpoints \(V_t\) and feasible traversal edges \(E_t\).  Using this structured representation, the robot incrementally selects a node as the next viewpoint. Since decisions are made upon arriving at the last selected viewpoint, the trajectory naturally forms as a sequence of waypoints \(\tau_i \in V\).

\begin{figure}[thpb]
  \centering
  \includegraphics[scale=0.5]{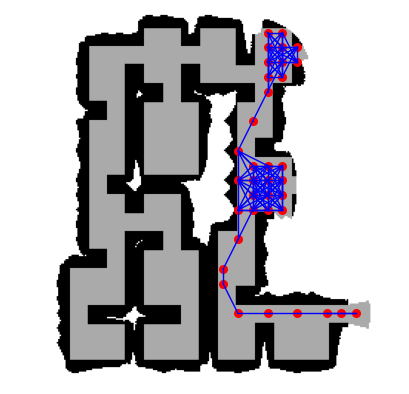}
  \caption{sparse information graph.}
  \label{partial_map}
\end{figure}

For improved decision-making, we compute node-specific information based on the environment. Here, we use the utility \(u_i\) at node \(v_i\) as the node-specific information. It is defined as the number of observable frontiers, representing potential unexplored areas within the line of sight. A frontier is considered observable if the direct line \(L(v_i, f_j)\) between node \(v_i\) and frontier \(f_j\) is collision-free and shorter than the sensor range \(d_s\). The utility at \(v_i\) is formulated as:

\begin{equation}
\begin{aligned}
    &u_i = |F_{o, i}|, \\
    &\forall f_j \in F_{o, i}, \quad \| f_j - v_i \| \leq d_s, \\
    &L(v_i, f_j) \cap (\mathcal{M} - \mathcal{M}_f) = \emptyset.
\end{aligned}
\end{equation}

where \(F_{o, i}\) denotes the set of observable frontiers at node \(v_i\). As shown in Fig.\ref{partial_map}, nodes with \( u_i = 0 \) (non-informative regions) are pruned, resulting in a sparse information graph \( G^s = (V^s, E^s) \) as the robot belief, where $V^s=\left\{v_i \mid u_i>0\right\}$ and $E^s \subseteq E_t$ preserves connectivity.

\subsection{Constructing Mixed Observation with Prompts}
\begin{tcolorbox}[fonttitle = \small\bfseries, title=Graph Describer Template,colframe=gray!2!black,colback=gray!2!white,boxrule=1pt,boxsep=0pt,left=5pt,right=5pt,fontupper=\footnotesize, halign title = flush center]

Currently, a robot is at coordinates ($\color[RGB]{255,190,122}{x_0},\color[RGB]{255,190,122}{y_0}$). 

Here is the sparse graph information for better decision making:

Nodes in sparse graph:

\#sparse node status

node i: Coordinates: ($\color[RGB]{216,56,58}{x_i},\color[RGB]{216,56,58}{y_i}$), {\color[RGB]{130,176,210}{$<$node-specific information$>$}},

...

Edges in sparse graph:

\#edges status

({\color[RGB]{142,207,201}{node\_index m}}, {\color[RGB]{142,207,201}{node\_index n}}), {\color[RGB]{130,176,210}{$<$edge information$>$}},

...
\end{tcolorbox}
\textbf{Graph Describer}: The graph describer \(D\) is responsible for generating task-agnostic textual descriptions of a given graph \(G\). To ensure the clarity and accuracy of these descriptions, we develop a graph-describing template. This template is designed to accommodate a wide range of graph configurations, including both directed and undirected graphs, as well as those with or without node and edge weights. To generate a description for a specific graph, the template replaces placeholders with actual data, such as the number of nodes, edges, and the endpoints of each edge. This allows for the creation of detailed and tailored descriptions that accurately represent the structure of the graph in question. The process of generating the textual description is formulated as \(D_G = D(G)\), where \(D_G\) denotes the description produced by the graph describer.  

By leveraging this unified and structured template, the graph describer can consistently produce coherent and informative descriptions that highlight the fundamental characteristics of the graph, independent of the particular task at hand.

\begin{tcolorbox}[fonttitle = \small\bfseries, title=Task Prompts Questioner Template,colframe=gray!2!black,colback=gray!2!white,boxrule=1pt,boxsep=0pt,left=5pt,right=5pt,fontupper=\footnotesize, halign title = flush center]

A robot is {\color[RGB]{170, 193, 240}\textbf{$<$Task Overview$>$}}.

{\color[RGB]{213, 232, 212}\textbf{$<$Graph Description$>$}} 

Preference:

{\color[RGB]{170, 193, 240}\textbf{$<$Preference Description$>$}}

{\color[RGB]{255, 229, 153}\textbf{$<$Task Intermediate Memory Description$>$}}

Based on the current state, which node should the robot move to next? 

Output Format:

{\color[RGB]{248, 206, 204}\textbf{$<$Output Format Description$>$}}

\end{tcolorbox}

\textbf{Task Prompts Questioner}: The questioner $Q$ is tailored to capture the intricate requirements of specific preference and reflect them in its output human-specific query. In detail, $Q$ receives the task-agnostic textual descriptions from the graph describer and refines them to align with the task context by elucidating the concrete meanings of nodes and edges. These refined descriptions are then enriched with task responsibilities and input/output specifications to form task-specific queries. The formulation of the questioner to generate the task-specific queries can be represented as follows:

\begin{equation}
Q_G^{H,I}=Q\left(D_G,H,I\right)
\end{equation}

where $Q_G^{H,I}$ represents the human-prefered task query generated by the questioner with given the preference $H$, intermediate memory $I$ and the textual description $D_G$. 

The construction of human-preferred queries can be approached in two primary ways: manual template-based construction and bootstrapping LLM agents. Manual template-based construction enhances \(D_G\) with human-preferred templates, which is ideal for tasks that have precise and well-defined requirements. For example, tasks like solving the Traveling Salesman Problem (TSP) require fine-grained operations on graphs, where accuracy is crucial and the task is well-understood. This method ensures clarity and minimizes errors by providing meticulous attention to detail. In contrast, bootstrapping LLM agents for automated synthesis offers a more flexible and cost-effective solution, especially for dynamic or bespoke tasks with vague or subjective preferences. These tasks, where precise requirements are not clearly defined, benefit from the adaptability of LLM agents. By interpreting context and generating queries autonomously, LLM agents reduce the manual effort required while enhancing the system’s ability to respond to changing conditions.

\subsection{Sequential Decision-making on a Graph}

Given that the free area is updated as the robot moves, online planning for autonomous exploration inherently becomes a sequential decision-making problem, which aligns well with the autoregressive architecture of LLMs. Therefore, we adopt an LLM-based planner for this task. We use the prompt constructed from \(Q_G^{H,I}\) as the planner's mixed observation \(o_t\). Consider the graph \(G^s\), which extends incrementally during exploration as the free areas are updated. At any given time, the robot’s current position is \(p_t\). Our LLM-based planner makes decisions on the next waypoint, selecting a node \(v_i\) such that \(\pi(a_i \mid o_t) = p(v_i \mid o_t)\), where \((p_t, v_i) \in E\) and the robot navigates toward the selected waypoint. Importantly, the planner only makes decisions upon reaching the previously selected waypoint, allowing for fixed-frequency decision-making to ensure a more reactive response to real-time map updates. One of the decision-making processes is as follows:

\begin{tcolorbox}[fonttitle = \small\bfseries, title=Decision Step Example,colframe=gray!2!black,colback=gray!2!white,boxrule=1pt,boxsep=0pt,left=5pt,right=5pt,fontupper=\footnotesize, halign title = flush center]

Okay, let's see. The robot is currently at [-4.0, 8.0], which is node 0. The next move should prioritize exploring unknown areas, especially in the upper-left corner. Also, avoid revisiting nodes unless necessary.
Looking at the edges connected to the current node (node 0), the possible next nodes are: (0.0, 16.0) node3, (-4.0, 16.0) node1, (0.0, 12.0) node5, (-4.0, 12.0) node2, (0.0, 8.0) node4, and itself. Since the robot's path history only includes the current node, we can ignore the self-loop.

Now, check the utilities of these nodes. Node1 (33), node3 (32), node5 (20), node2 (28), node4 (7). The highest utility here is node1 and node3. But node1 is at (-4.0,16.0), which is up along the same x-axis, so upper-left. Node3 is (0.0,16.0), which is more to the right. Since the preference is upper-left, node1 is better. Also, node1's utility is slightly higher (33 vs 32). So moving to node1 makes sense.

Wait, but what's the distance? The current node is at (-4,8). The distance to node1 (-4,16) is sqrt(0\textasciicircum2 +8\textasciicircum2)=8 units. To node3 (0,16), it's sqrt(4\textasciicircum2 +8\textasciicircum2)=sqrt(16+64)=sqrt(80)$\approx$8.94. So node1 is closer. So node1 has higher utility and is closer, and is in the upper-left. So definitely node1 is the best choice. The other options like node2 (28) is lower utility. Node5 (20) is even lower. Node4 is very low. So the robot should move to node1. There's no revisiting here since the path history is only step 0. So the answer should be [-4.0, 16.0].
\end{tcolorbox}

\begin{figure*}[t]
  \centering
  \begin{subfigure}{0.42\textwidth}
    \centering
    \includegraphics[width=\textwidth]{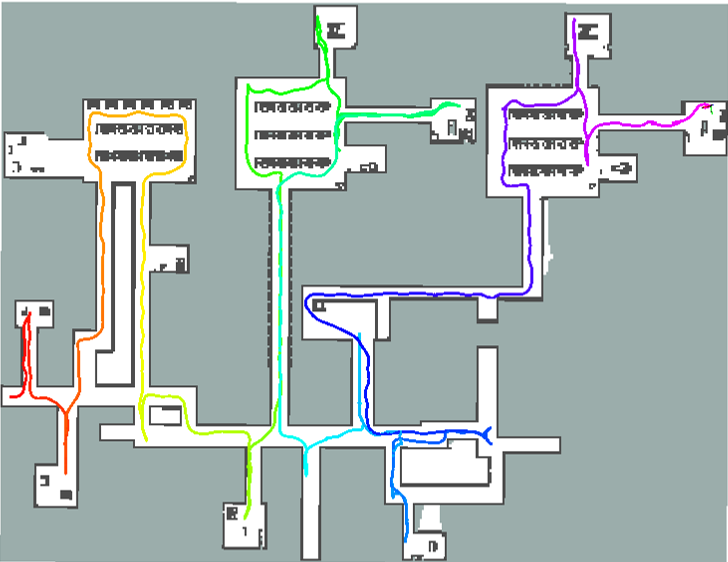}
    \caption{HELM\_PRE, distance 913.1$m$}
    \label{fig:gahelm_pre}
  \end{subfigure}
  \hspace{0.05\textwidth} 
  \begin{subfigure}{0.4\textwidth}
    \centering
    \includegraphics[width=\textwidth]{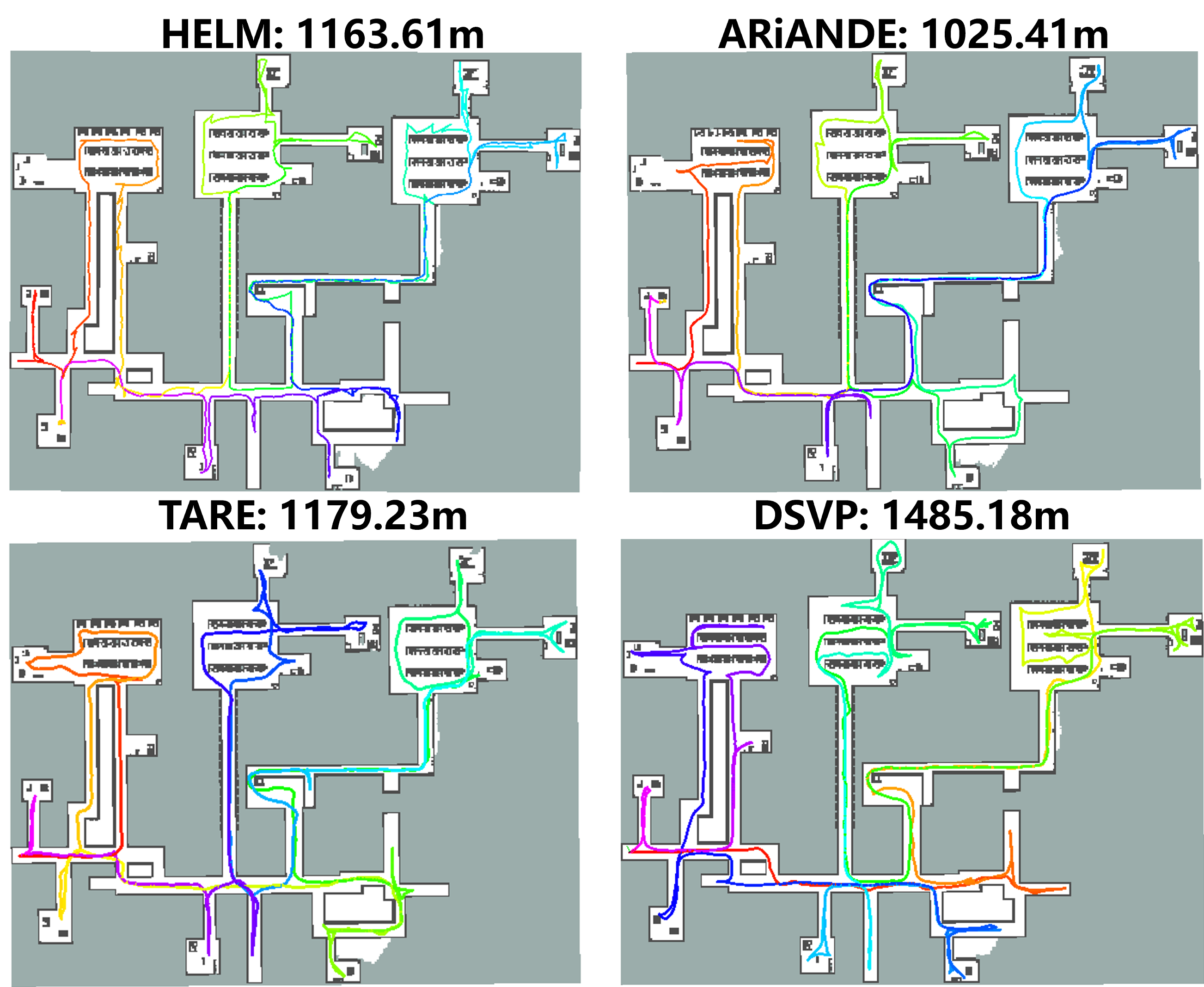}
    \caption{Other Baselines}
    \label{fig:garl}
  \end{subfigure}

  \caption{Exploration paths comparisons in a large-scale $130m \times 100m$ indoor office simulation.}
  \label{gazebo_simu}
\end{figure*}

\begin{table*}[t]
\vspace{+0.15cm}
\caption{
\textbf{Comparisons with baseline planners in dungeon environments (identical 100 scenarios for each method).}}
\label{table:1}
\vspace{-0.2cm}
\begin{center}
\begin{tabular}{c|c|c|c|c|c|c|c}
\toprule
& Nearest & NBVP & TARE Local & ARiADNE & DARE & Ours & Optimal\\
\midrule
Distance (m) & 578.49($\pm$65.01)& 673.46($\pm$119.79) & 552.44($\pm$ 60.43) & 557.35($\pm$76.38) & 572.08($\pm$63.19) & 557.66($\pm$65.01) &  483.04($\pm$63.33)\\
\midrule
Gap to Optimal & 19.7\% & 39.4\% & 14.4\% & 15.4\% & 18.4\% & 15.4\% & 0\%\\
\bottomrule
\end{tabular}
\end{center}
\vspace{-0.5cm}
\end{table*}

\subsection{Closed-Loop Robot Control}
After receiving the decision output, we extract the current waypoint, which is selected from the graph we constructed. Using this waypoint, we compute a connected shortest path. This path is then sent to the robot, directing it to sequentially visit each waypoint. The path generation is handled by the LLM-agent, which first determines the most suitable algorithm to use and generates the corresponding code for efficient reuse. Once the robot has visited all the waypoints along the path, the closed-loop robot control process is complete.

\section{EXPERIMENTS}

\subsection{Comparison Analysis}

To evaluate the fundamental exploration capability of our planner, we test   HELM without additional human preferences—only the basic task requirements are provided. To assess its performance, we conduct comparison experiments using a test set consisting of 100 unseen simulated dungeon environments. We compare the travel distance of HELM to complete the exploration with several conventional planners: (1) \textbf{Nearest} \cite{1997frontier}, where the robot always moves to the closest frontier; (2) \textbf{NBVP} \cite{nbvp}, a sampling-based planner that uses rapid random trees to determine the optimal viewpoint; (3) \textbf{TARE Local} \cite{tare}, the local-level planner of TARE, which plans the shortest trajectory covering all frontiers in the current partial map; (4) \textbf{ARiADNE} \cite{ariadne}, a state-of-the-art graph-based DRL planner; (5) \textbf{DARE} \cite{dare}, a generative diffusion-based DRL planner; (6) \textbf{Optimal}, the shortest path for covering a known environment directly calculated based on the ground truth. Specifically, we construct a ground truth graph \( \mathcal{M}_g(V^*, E^*) \), a collision-free graph that covers the free area \( M_f \). The boundary of the unexplored free area \( (\mathcal{M}_f - \mathcal{B}_f) \) is considered as the ground truth frontiers to be observed. We then sample nodes such that visiting these nodes allows the robot to cover all ground truth frontiers. The shortest path to visit these nodes is found by solving the TSP. This sampling process repeats \( k \) times, and the shortest path from the samples is selected as the expert demonstration.



\begin{figure*}[t]
  \centering
  \begin{subfigure}{0.45\textwidth}
    \centering
    \includegraphics[width=\textwidth]{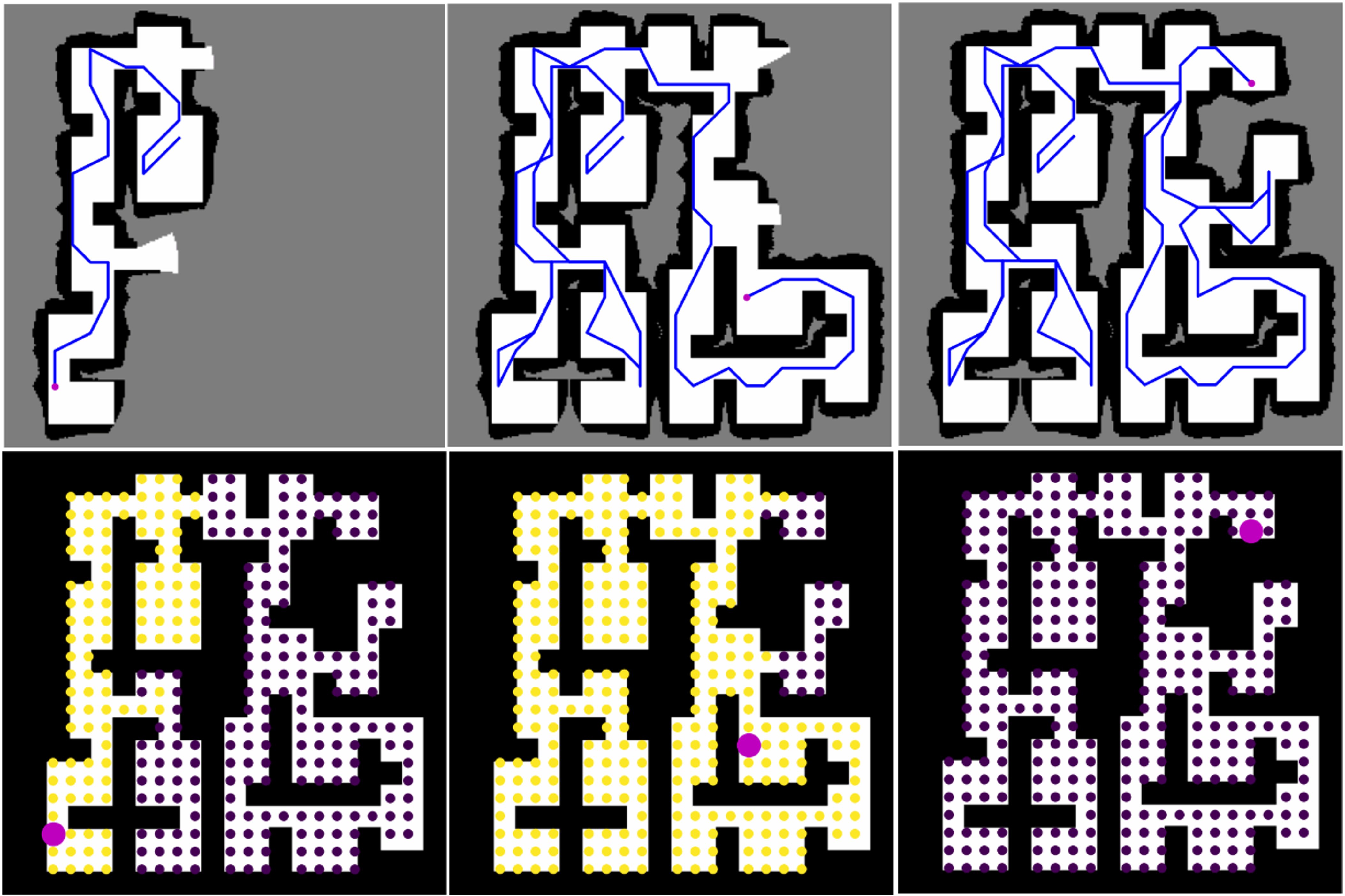}
    \caption{The robot prioritized exploring the bottom-left corner.}
    \label{fig:prompt1}
  \end{subfigure}
  \hspace{0.03\textwidth} 
  \begin{subfigure}{0.45\textwidth}
    \centering
    \includegraphics[width=\textwidth]{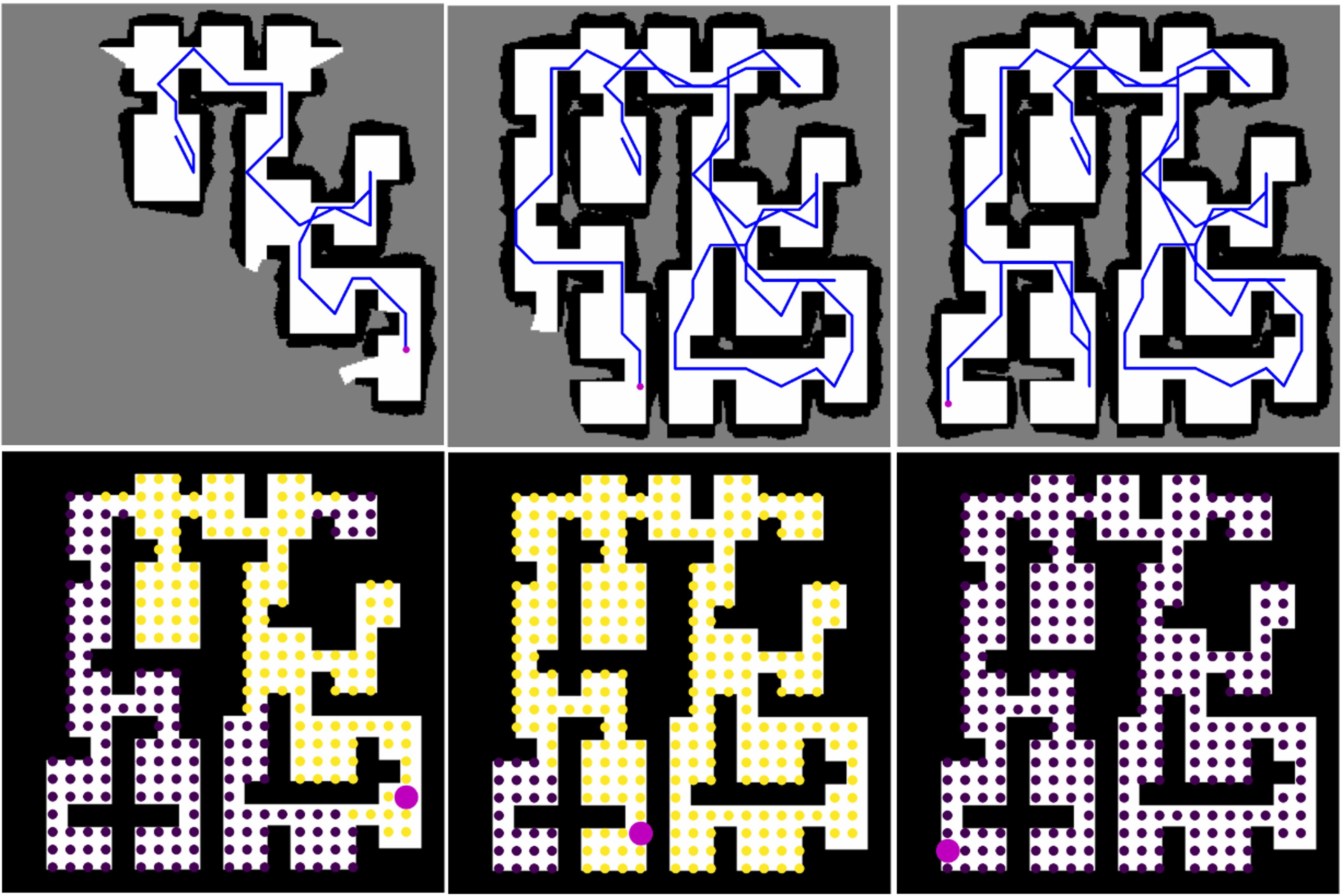}
    \caption{The robot prioritized exploring the bottom-right corner.}
    \label{fig:prompt2}
  \end{subfigure}
  \caption{Different human preferences lead to different exploration strategies. The upper part of the image shows the robot's trajectory and the partial map obtained during the exploration process. The lower part shows the ground truth, with yellow representing the explored areas and purple representing the unexplored areas.}
  \label{prompt_exp}
\end{figure*}

In our experiments, exploration is considered complete once more than 99\% of the free area has been covered, allowing for minor tolerances in error. The mean and variance of the trajectory lengths required for full exploration are summarized in Table \ref{table:1}. We use Deepseek-V3 as the foundation LLM \cite{deepseekv3},  the results show that HELM, leveraging NLP to provide human-tuned exploration behaviors, achieves competitive efficiency in terms of travel distance compared to other advanced conventional and learning-based baselines while performing on par with TARE Local, the state-of-the-art planner.  Moreover, we note that our method is training-free, which significantly enhances its generalization ability. This allows HELM to quickly adapt to different task scenarios without extensive retraining, providing greater flexibility and scalability across diverse environments.

\subsection{Gazebo Simulation}
\begin{figure}[thpb]
  \centering
  \hspace{-0.05\textwidth}
  \includegraphics[scale=0.25]{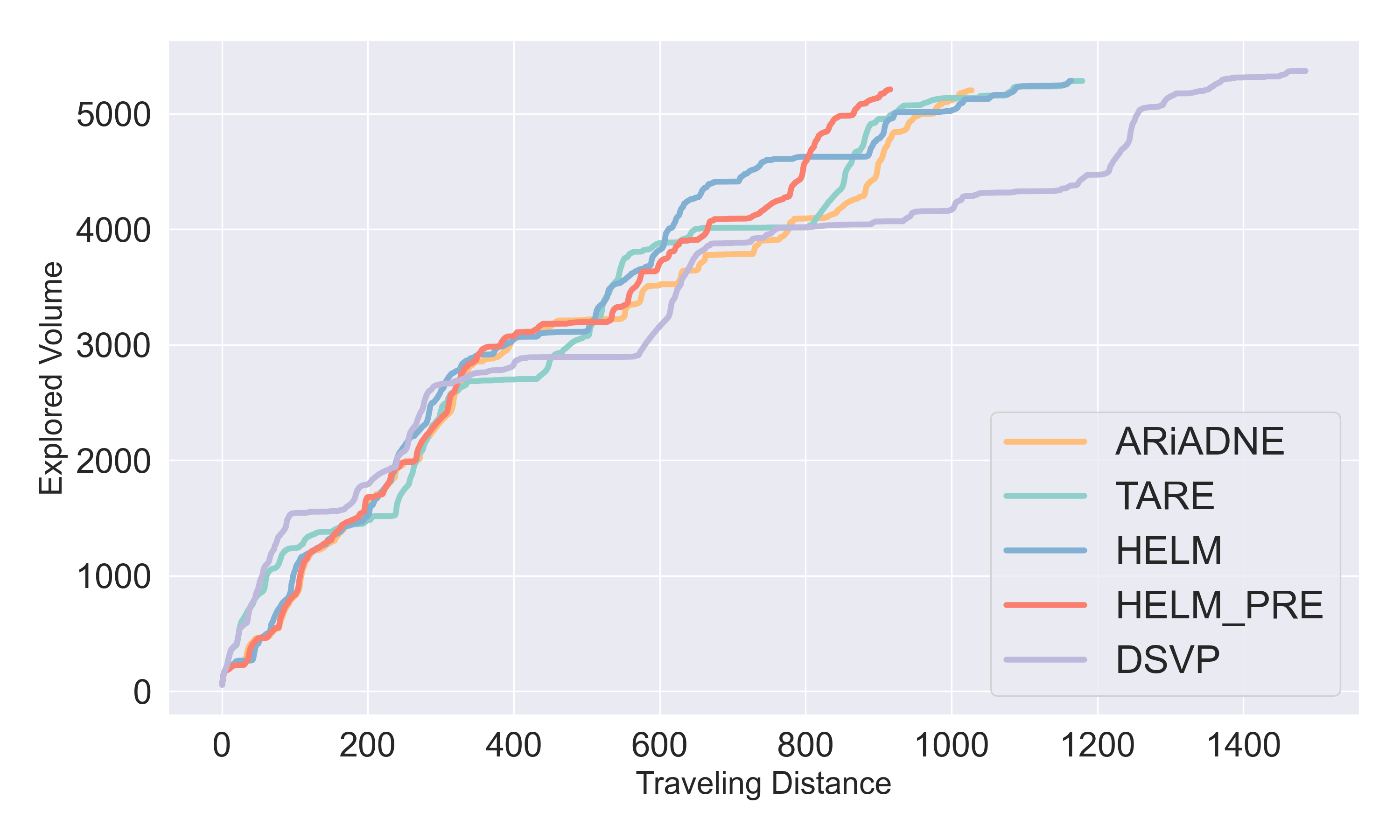}
  \caption{Trajectory analysis in the Gazebo simulation.}
  \label{exptravel}
\end{figure}

We implement HELM in ROS and perform a high-fidelity simulation in a large-scale $130m \times 100m$ indoor office environment, provided by \cite{tareenv}. The Gazebo experimental results indicate that incorporating human preferences into HELM can significantly improve its performance. The simulation incorporates a real-world sensor model (Velodyne 16-line 3D LiDAR) and robot constraints (four-wheel differential drive with a maximum speed of 2m/s). Using Octomap, we classify free and occupied areas to define the "explored" region, with a map resolution of 0.4m, a sensor range of 20m, and a node resolution of 2m. 

The results for Traveling Distance and Explored Volume are shown in Fig.\ref{exptravel} and the exploration trajectories are shown in Fig.\ref{gazebo_simu}. The travel distances and replanning time are as follows: HELM\_PRE: 913.1m, 0.2s; HELM: 1163.61m, 0.2s; ARiADNE: 1025.41m, 0.2s; TARE: 1179.23m, 0.3s and DSVP: 1485.18m, 0.3s. HELM's preference is set as \textbf{``Prioritize Unknown Areas: Focus on unexplored regions, Avoid revisiting the history path nodes''}, while HELM\_PRE incorporates an additional preference \textbf{``Prioritize Unknown Areas: Focus on unexplored regions, with the highest priority given to the left half of the map, ensuring it is fully explored before moving to the right half. Within the left half, prioritize the frontiers of the upper-left and lower-left regions''}, As shown in Fig.\ref{fig:gahelm_pre}, the trajectory of HELM\_PRE clearly prioritizes exploration of the left half without returning, whereas all other baselines revisit the left half during exploration. This demonstrates that HELM can understand and leverage natural language preferences to assist in decision-making. Notably, without additional preferences, HELM performs on par with TARE, while with the extra preference, HELM achieves state-of-the-art performance, surpassing all baselines. This also highlights that HELM maintains strong generalization even in large-scale environments.

\subsection{Preference Experiment}

To evaluate the effectiveness of HELM in accommodating different human preferences, we conducted experiments on a larger dungeon map, as shown in Fig.\ref{prompt_exp}. In Fig.\ref{fig:prompt1}, the human preference provided was: \textbf{``The robot should focus on exploring the unexplored regions, particularly those in the bottom-left corner of the map."} In Fig.\ref{fig:prompt2}, the preference was: \textbf{``The robot should prioritize investigating the unknown areas, especially those in the bottom-right corner of the map."} While the preferences differ, all other experimental conditions remained unchanged. As seen in Fig.\ref{fig:prompt1}, the robot prioritizes the bottom-left corner, aligning with the specified preference, whereas in Fig.\ref{fig:prompt2}, it prioritizes the bottom-right corner. After fulfilling the preference-driven goal, the robot continues its exploration until the task is completed. 

\section{CONCLUSIONS}
In this work, we propose HELM, a novel framework for learning-based autonomous exploration that relies on an LLM to integrate human preference during the mission. Our experiments demonstrate that HELM achieves on-par performance with state-of-the-art exploration planners while providing significantly enhanced flexibility in human-robot interaction. The framework's ability to seamlessly incorporate human intent allows for more efficient and user-friendly exploration in dynamic environments. 

In future work, we plan to extend HELM to support a wider variety of environments, tasks, and multi-robot systems. Additionally, we aim to advance autonomous exploration towards vision-based semantic scene understanding and reconstruction, enabling robots to not only navigate unknown environments but also interpret and interact with them in a more intelligent and meaningful way.










\end{document}


\appendices
\section*{Appendex}
\renewcommand{\thesection}{\Alph{section}}
\subsection*{A. Proof Of The Global Section Condition}

Let $\mathcal{F}$ be a sheaf on a topological space $X$, and consider two vertices $u$ and $v$ connected by an edge $e$. Assume $U_u$ and $U_v$ are neighborhoods of $u$ and $v$, respectively, covering the edge $e$.

Define local sections $x_u \in \mathcal{F}(U_u)$ and $x_v \in \mathcal{F}(U_v)$. The sheaf's compatibility condition requires:
\begin{equation}
    x_u|_{U_u \cap U_v} = x_v|_{U_u \cap U_v}
\end{equation}
on the overlap $U_u \cap U_v$, which includes the edge $e$.

We focus on the edge $e \subseteq U_u \cap U_v$. By the compatibility requirement:
\begin{equation}
    \mathcal{F}_{v \unlhd e} x_v = \mathcal{F}_{u \unlhd e} x_u
\end{equation} 

where $\mathcal{F}_{v \unlhd e}$ and $\mathcal{F}_{u \unlhd e}$ represent the restrictions of $x_v$ and $x_u$ to the edge $e$.

The equality $\mathcal{F}_{v \unlhd e} x_v = \mathcal{F}_{u \unlhd e} x_u$ confirms that the local sections agree on their overlap on $e$, thereby satisfying the sheaf condition. This condition guarantees that local data can be consistently extended to global data across overlaps, fundamental to the structure of a sheaf.

\subsection*{B. Performance Metric}
\textbf{Episode Length (EL)}: Used to measure the optimality of the solution produced by the planner. A shorter episode length means that all agents can reach the goal faster.

$$
e l=\sum_{i=1}^N \frac{p l_i}{N}
$$

$p l_i$ represents the path length the last agent requires to reach the goal when all agents in the $i$-th episode reach the goal.

\textbf{Success rate (SR)}: It is used to measure the ability of the planner to complete a MAPF task, and the following formula can express its calculation:

$$
s r=\frac{s}{N}
$$

where $s$ represents the number of episodes in which all agents reach the goal within the specified number of steps, and $N$ is the total number of episodes tested.

\textbf{Arrival Rate (AR)}: It is used to measure the ability of the planner to complete lifelong tasks. A higher arrival rate means that the planner has better potential for lifelong MAPF tasks.

$$
a r=\sum_{i=1}^N \frac{a r r_i}{N \times n}
$$

Among them, $n$ represents the number of agents in each episode, and $a r r_i$ represents the number of agents reaching the goal in the $i$-th episode.

\subsection*{C. Section Loss}

We introduced a section loss and incorporated it into the Q-value calculation, effectively guiding agents towards consensus. The test results following training, as shown in Table \ref{loss}, indicate that the section loss values tend to be very small (ranging from $10^{-4}$ to $10^{-3}$).

\begin{table}[h]
\centering
\caption{section loss}
{\fontsize{7.5pt}{12pt}\selectfont
\setlength{\tabcolsep}{4pt}
\begin{tabular}{c|cccccc}
\toprule
\diagbox[dir=NW]{\textbf{Map}}{\textbf{Agent}} & \textbf{4} & \textbf{8} & \textbf{16} & \textbf{32} & \textbf{64} & \textbf{128} \\
\midrule
40*40 & 6.63e-04 & 8.83e-04 & 1.37e-03 & 9.31e-04 & 4.30e-04 & 1.08e-04 \\
60*60 & 3.41e-04 & 5.96e-04 & 9.62e-04 & 1.13e-03 & 9.51e-04 & 4.74e-04 \\
\bottomrule
\end{tabular}
}
\label{loss}
\end{table}

\subsection*{D. Hyperparameters}
Table \ref{hyperparameters} describes the primary hyperparameters in our method. The computational simulations and experiments for our study were executed on Ubuntu 20.04 equipped with an AMD EPYC 7T83 CPU and a Nvidia GeForce RTX 3090ti GPU. As for the software framework, the experiments leveraged Python version 3.8.18 coupled with PyTorch version 2.0.0 to ensure efficient computation and reproducibility.
\begin{table}[h]
\centering
\caption{Hyperparameters table}
\begin{tabular}{@{}ll@{}}
\toprule
\textbf{Hyperparameters} & \textbf{Value} \\ \midrule
Number of agents& 5\\
FOV size& 9\\
Maximum episode length& 512\\
World size& (10,40)\\
Capacity of episodic buffer& 2048\\
gamma & 0.9 \\
batch size& 192\\
optimize & Adam \\
prioritized replay $\alpha$& 0.6\\
prioritized replay $\beta$& 0.4\\ \bottomrule
\end{tabular}
\label{hyperparameters}
\end{table}